\documentclass[11pt]{article} 
\usepackage{fullpage}
\usepackage{hyperref}
\usepackage{url}
\usepackage{amsfonts}
\usepackage{amsmath}
\usepackage{color}
\usepackage{graphicx}
\usepackage{amstext}
\usepackage{url}
\usepackage[boxed]{algorithm2e}

\newtheorem{theorem}{Theorem}

\def\cost{\text{cost}}
\def\nbop{\text{\#ops}}

\title{Further heuristics for $k$-means: The merge-and-split heuristic and the $(k,l)$-means}

\author{
Frank Nielsen\\
Sony Computer Science Laboratories, Japan\\
\'Ecole Polytechnique, France\\
\texttt{Frank.Nielsen@acm.org} \\
\and
Richard Nock\\
NICTA, Australia\\
\texttt{Richard.Nock@nicta.com.au}
}

\date{\ }

\def\bbN{\mathbb{N}}
\def\calX{\mathcal{X}}

\def\bbR{\mathbb{R}}
\def\bbN{\mathbb{N}}

\def\calK{\mathcal{K}}
\def\calC{\mathcal{C}}
\def\calX{\mathcal{X}}
\def\calP{\mathcal{P}}
\def\bbR{\mathbb{R}}

\def\NN{\mathrm{NN}}
\def\iNN{\mathrm{iNN}}
\def\bbN{\mathbb{N}}

\def\maxIter{\mathrm{maxIter}}
\def\Iter{\mathrm{Iter}}

\def\mstart{\mathrm{mstart}}
\def\CH{\mathrm{CH}}

\def\cost{\mathrm{cost}}
\def\calX{\mathcal{X}}
\def\calC{\mathcal{C}}

\newtheorem{fact}{Fact}

\begin{document}

\maketitle

\begin{abstract}
The $k$-means clustering problem asks to partition the data into $k$ clusters so as to minimize the sum of the squared Euclidean distances of the data points to their closest cluster center.
Finding the optimal $k$-means clustering of a $d$-dimensional data set is NP-hard in general and many heuristics have been designed for minimizing monotonically  
the $k$-means objective function.
Those heuristics got trapped into local minima and thus heavily depend on the initial seeding of the cluster centers.  
The celebrated $k$-means++ algorithm is such a randomized seeding method which guarantees probabilistically a good initialization with respect to the global minimum.
In this paper, we first show how to extend Lloyd's batched relocation heuristic and Hartigan's single-point relocation heuristic to take into account empty-cluster and single-point cluster events, respectively. 
Those events tend to increasingly occur when $k$ or $d$ increases, or when performing several restarts of the $k$-means heuristic with a different seeding  at each round in order to keep the best clustering in the lot.
We show that those special events are a blessing because they allow to partially re-seed some cluster centers while further  
minimizing the $k$-means objective function. 
Second, we describe a novel heuristic, called merge-and-split $k$-means, that consists in merging two clusters and splitting this merged cluster again with two new centers provided it improves the $k$-means objective. 
Hartigan's heuristic can improve a Lloyd's heuristic when it reaches a local minimum, and similarly this novel heuristic can improve Hartigan's $k$-means when it has converged to a local minimum.
We show empirically that  this merge-and-split $k$-means improves over the Hartigan's heuristic which is the {\em de facto} method of choice. 
Finally, we propose the $(k,l)$-means objective that generalizes the $k$-means objective by associating the data points to their $l$ closest cluster centers, and show how to either directly convert or  iteratively relax the $(k,l)$-means into a $k$-means in order to reach better local minima.
\end{abstract}

\section{Introduction}

Clustering is the task that consists in grouping data into homogeneous clusters with the goal that intra-cluster data should be more similar than  inter-cluster data.
Let $\calP=\{p_1, ..., p_n\}$ be a set of $n$ points\footnote{For the sake of clarity and without loss of generality, we do not consider weighted points.} in $\bbR^d$.
Let  $\calC_1, ..., \calC_k$ be the $k$ {\em non-empty} clusters partitioning $\calP$ and denote by $\calK=\{c_1, ..., c_k\}$ the set of $k$ cluster centers, the {\em cluster prototypes}.
$k$-Means is one of the oldest  and yet prevalent clustering technique that consists in minimizing:

\begin{equation}\label{eq:kmeanscost}
e(\calP,\calK)=\sum_{i=1}^n \min_{j=1}^k D(p_i,c_j) = \sum_{i=1}^n  D(p_i,c_{l_i})= \sum_{j=1}^k \sum_{p\in\calK_j} D(p,c_j),
\end{equation}
where $D(p,q)=\|p-q\|^2$ denotes the {\em squared} Euclidean distance, and $l_i$ the index (or label) of the center of $\calK$ that is the closest nearest neighbor to $p_i$ (say, in case of ties, choose the minimum integer).
Finding an optimal clustering minimizing globally $\min_\calK e(\calP,\calK)$ is NP-hard when $d>1$ 
and $k>1$~\cite{kmeansNPhard-2009,twomeans-nphard-2007}, and polynomial when $d=1$ using dynamic programming~\cite{ClusteringDynamicProg-1973} or when $k=1$ setting $c$ to the center of mass.
Note that there is an {\em exponential number} of optimal $k$-means clustering yielding  the same optimal objective function: Indeed, consider an equilateral triangle with $n=3$ and $k=2$, we thus get $3$ equivalent optimal clustering related by rotational symmetries. Then make $s$ far away separated copies so that $n=3s$ and consider $k=2s$, we end up with $3^s=3^{\frac{n}{2}}$ optimal $k$-means clustering. Minimizing the $k$-means function of Eq.~\ref{eq:kmeanscost} is equivalent to minimizing the sum of intra-cluster squared distances or maximizing the sum of inter-cluster squared  distances:
\begin{equation}\label{eq:kmeanscostequiv}
\min_{\calK} e(\calP,\calK)  \equiv \min_{\calK} \sum_{j=1}^k  \sum_{p_i,p_j\in\calC_j} \|p_i-p_j\|^2  \equiv \max_{\calK} \sum_{j=1}^k \sum_{p_i\in\calC_j, p_j\not\in\calC_j} \|p_i-p_j\|^2
\end{equation} 

Many heuristics have been proposed to overcome the NP-hardness of $k$-means.
They can be classified into two main groups: The {\em local search} heuristics and the {\em global} heuristics that can be used to initialize the local heuristics.
For example, the following four heuristics are classically\footnote{See for example the R language for statistical computing, \url{http://www.r-project.org/}} implemented:
\begin{itemize}
	\item Forgy~\cite{Forgy-1965} (random): Draw uniformly at random $k$ points from $\calP$ to set the cluster prototypes $\calK$ inducing the partition.
	It can be proved that this best {\em discrete} $k$-means (with $\calK\subset\calP$) yields a $2$-approximation factor compared to the ordinary $k$-means using a proof by contradiction based on the {\em variance-bias decomposition}: $e(\calP,c') = v(\calP) + n \|c'-c\|^2$,
where $v(\calX)=\sum_{i=1}^n \|p_i - c\|^2 = \sum_{i=1}^n \|p_i\|^2 - n\|c\|^2$ denotes the variance and $c=\frac{1}{n}\sum_{i=1}^n p_i$ the centroid. In fact, $e(\calP,\calK)=\sum_{j=1}^k  v(\calC_j)$, the sum of intra-cluster variances (and $e(\calP,\calK)=\sum_{i=1}^n \|p_i\|^2 - \sum_{j=1}^k n_j \|c_j\|^2$).

	\item MacQueen~\cite{MacQueen-1967} (online): From a given initialization of the $k$ centers defining singleton clusters (say, $\calC_j=\{p_j\}$ for the $k$ clusters), we add a new point at a time to the cluster that contains the closest center, update that cluster centroid, and reiterate until convergence.
	This heuristic is also called the online or single-point $k$-means~\cite{Lloydkmeansiterations-2005}.
	
	\item Lloyd~\cite{Lloyd-1957} (batched): From a given initialization of cluster prototypes, (1) assign points to their closest cluster,  
	(2) relocate cluster centers to their cluster centroids, and reiterate those two steps until convergence.  
	
	\item Hartigan~\cite{Hartigan-1975,KmeansHartiganWong-1979} (single-point relocation): From a given initialization, find how to move a point from a cluster to another cluster so that the $k$-means cost of Eq.~\ref{eq:kmeanscost} strictly decreases and reiterate those single-point relocations until convergence is reached. Note that a point maybe assigned to a cluster which center is {\em not} its closest center~\cite{Hartigan-2010}. 
\end{itemize}

In general, a $k$-means clustering technique partitions the data into pairwise non-overlapping  convex hulls $\CH(\calC_1), ..., \CH(\calC_k)$: The {\em Voronoi partition}. A partition is said {\em stable} when a local improvement of the heuristic cannot improve its $k$-means score.
Let $P_{F,Q,L,H}(n,k)$ denotes the maximum number of stable $k$-means partitions obtained by Forgy's, MacQueen's, Lloyd's and Hartigan's schemes, respectively.

\begin{fact}[Voronoi partitions]
We have $P_F(n,k)\leq {n\choose k}$ and  $P_H(n,k) \leq P_L(n,k) \leq P_{\CH}(n,k) << P(n,k)$,
where $P(n,k)=\frac{1}{k!} \sum_{i=0}^k (-1)^{k-i} {k\choose i} i^n$ denotes the number of partitions of $n$ elements into $k$ non-empty subsets
 (that is, the Stirling numbers of the second kind) and $P_{\CH}(n,k)$ denotes the number of partitions with non-overlapping (and non-empty) convex hulls (that is, the number of $k$-Voronoi partitions).
\end{fact}
Hartigan's single-point relocation heuristic may improved Lloyd's clustering but not the converse~\cite{Hartigan-2013}.
Note that Lloyd's heuristic may require an exponential number of iterations to converge~\cite{kmeansexponential-2011}.
It is an open question~\cite{Hartigan-2010} to bound the maximum number of Hartigan's iterations.

On one hand, for those local heuristics performing pivots on Voronoi partitions using primitives, initialization (i.e., the initial Voronoi partition) is crucial~\cite{InitializationKmeans-2012} to obtain a good clustering, and several restarts, denoted by $\mstart$, are performed in practice to choose the best clustering.
In practice, Forgy's initialization has been replaced by $k$-means++~\cite{kmeanspp-2007} which provides an {\em expected} $\bar O(\log k)$  competitive initialization. However, it was shown  that there exits point sets (even in 2D) for which the probability to get such a good initialization is exponentially low~\cite{kmeanspp-2014}  (and thus requiring exponentially many initialization restarts to reach a good Voronoi partition with high probability). 

On the other hand, the {\em global $k$-means}~\cite{GlobalKMeans-2003,GlobalKmeans-2011} builds incrementally the clustering by adding one seed at a time. Given a current $s$-clustering it chooses the point in $\calP$ that minimizes the $(s+1)$-means objective function.
Thus initialization is limited to choosing the first point, and all points can be considered as this first starting point.
However, Global $k$-means requires more computation. 


In this paper, we do not address the problem of choosing the most appropriate number, $k$, of clusters: This model selection problem has been investigated in~\cite{Xmeans-2000,DP-kmeans-2012}. We also consider the squared Euclidean distance although the results apply to any other Bregman divergence~\cite{bregmankmeans-2005,Hartigan-2013}. 

The paper is organized as follows: We investigate the blessing of empty-cluster exceptions in Lloyd's heuristic in Section~\ref{sec:ece}, and of
single-point-cluster exceptions in Hartigan's scheme in Section~\ref{sec:spe}.
In Section~\ref{sec:msc}, we describe our novel heuristic merge-split-cluster $k$-means and report on its performances with respect to Hartigan's heuristic.
In Section~\ref{sec:klmeans}, we present a generalization of the $k$-means objective function where each point is associated to its $l$ closest clusters: the $(k,l)$-means clustering. We show how to directl convert or iteratively relax a sequence of $(k,l)$-means to a $k$-means and compare experimentally those solutions with a direct $k$-means. Finally, Section~\ref{sec:dis} wrap ups the contributions and discusses further perspectives.

\section{The blessing of empty-cluster exceptions in Lloyd's batched $k$-means\label{sec:ece}}
Lloyd's $k$-means~\cite{Lloyd-1957} starts by initializing the seeds of the cluster centers $\calK=\{c_1, ..., c_k\}$, and then iterates by assigning the data to their closest cluster center with respect to the squared Euclidean distance, and then relocates the cluster centers to their centroids.
Those batched assignment/relocation iterations are repeated until convergence is reached: The $k$-means cost monotonically decreases  with guaranteed convergence after a finite number of iterations~\cite{VarianceClustering-1994}. The complexity of Lloyd's $k$-means is $O(ndks)$ where $s$ denotes the number of iterations.
It has been proved that Lloyd's $k$-means performs a maximum number $s$ of iterations exponential~\cite{kmeansexponential-2011} or  polynomial in $n$, $d$ and the spread\footnote{The spread $\Delta$ is the ratio of the maximum point inter-distance over the minimum point inter-distance.} of the point set~\cite{kmeans-analysis-2000}. 
Some 1D point set  are reported to take $\Omega(n)$ iterations even for $k=2$, see~\cite{Lloydkmeansiterations-2005}.
We first, report a lower bound on the number of Lloyd's stable optima $P_L(n,k)$:

\begin{fact}[Exponentially many Lloyd's $k$-means minima]
Lloyd's $k$-means may have $P_L(n,k)=\Omega(2^{\frac{n}{2k}})$ stable local minima.
\end{fact}
The proof follows from the gadget illustrated in Figure~\ref{fig:kmeans}.

\def\ttt{0.15\columnwidth}
\begin{figure}[ht]%
\centering
(a)\fbox{\includegraphics[width=\ttt]{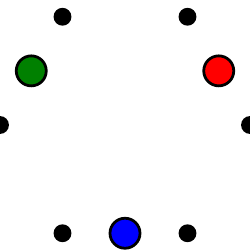}}
(b)\fbox{\includegraphics[width=\ttt]{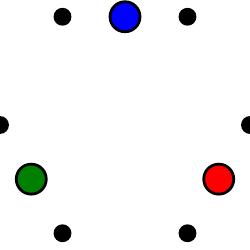}}
(c)\fbox{\includegraphics[width=\ttt]{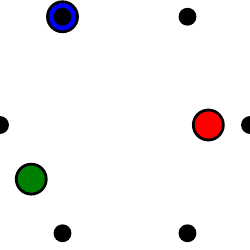}}
(d)\fbox{\includegraphics[width=\ttt]{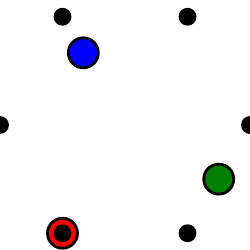}}\\ \def\ttt{0.15\columnwidth}
(a)\fbox{\includegraphics[width=\ttt]{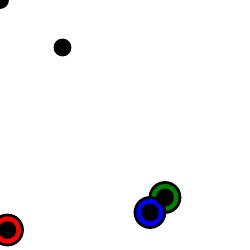}} 
(b)\fbox{\includegraphics[width=\ttt]{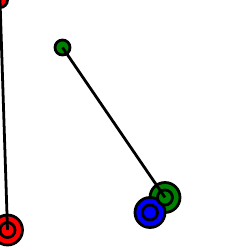}} 
(c)\fbox{\includegraphics[width=\ttt]{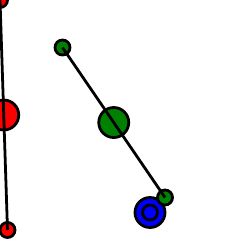}} 
(d)\fbox{\includegraphics[width=\ttt]{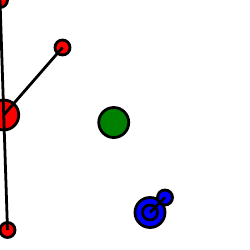}}  
 
\caption{Top: Lloyd' s $k$-means may have an exponential number of stable optima: Use locally the $k=2p$-gon (here $p=3$) gadget that admits $2$ global solution (a) and (b).
Lloyd's $k$-means can be trapped into a local minimum: Cost in (c) and (d) is $\sim 0.5417$ compared to the global minima $0.375$) in (a) and (b).
Centroids are depicted by large colored disks. Bottom: Lloyd's $k$-means local optimization technique may produce  {\em empty cluster exceptions}.
Consider $n=5$ points and $k=3$ clusters: $p_1=(0,0)$, $p_2=(0.25,0.19)$, $p_3=(0.03, 0.92)$, $p_4=(0.66,0.79)$ and $p_5=(0.6,0.85)$ with $k=3$ and
``random'' Forgy initialization: $c_1=p_3, c_2=p_4$ and $c_3=p_5$.
Then the initial $k$-means cost (a) is $1.3754$, the first iteration (b) and (c) yields cost $0.6877$ and then at the second iteration we have an empty cluster exception in (d): The green cluster. 
\label{fig:emptyexception}
\label{fig:kmeans}%
}
\end{figure}

\begin{algorithm}[H]
\SetAlgoLined
 \KwData{$\calP=\{(w_1,p_1), ..., (w_n,p_n)\}$ a data set of size $n$, $k\in\bbN$: number of clusters}
 \KwResult{A clustering partition $\calC_1, ...,\calC_k$ where each point belongs to exactly one cluster (hard membership)}
 \underline{Initialization}: Get $k$ cluster centers $\calC=\{c_1, ..., c_k\}$ by choosing cluster prototypes at random from $\calP$ (e.g., Forgy or $k$-means++)\;
 $\Iter \leftarrow 0$\;
 \While{not converged}{
 Increment $\Iter$, $e=0$\; 
  (a) \underline{Assign} each point $p_i$ to its closest cluster $\calC_{a_i}$\;
	\tcc{$\iNN$ denotes the index of the nearest neighbor}
	$$
	a_i=\iNN(p_i;\calK).
	$$
	(b) \underline{Relocate} each cluster prototype $c_j$ by taking the center of mass of its assigned points\;
	$$
	\calC_j=\{ p\in\calP : j=\iNN(p;\calC)\},\quad n_j=\sum_{p_l\in\calC_j} w_l.
	$$
 \eIf{$n_j>0$}
	{
	Non-empty cluster and centroid relocation:
	$$
	c_j= \frac{1}{n_j} \sum_{p_l\in\calC_j} w_lp_l
	$$}
	{
	$e \leftarrow e+1$\;
	}
	(c) \underline{New seeding} \;
	\tcc{Empty cluster exception (may have occured overall $\Omega(k)$ times)}
	\lnl{lloyd:exception} Choose $e$ new seeds for the empty clusters using $k$-means++ or global $k$-means, etc.\;
	Check for convergence by checking if at least one $a_i$ is different from the previous iteration\;
	\If{$\Iter>\maxIter$}{break\;}
 }
 \caption{Extended Lloyd's $k$-means clustering: batched updates handling empty cluster exceptions.\label{alg:lloyd}}
\end{algorithm}

The Hartigan's heuristic~\cite{Hartigan-1975,KmeansHartiganWong-1979} proceeds by relocating a  single point between two clusters provided that the $k$-means cost function decreases.
It  can thus further decrease the $k$-means score when Lloyd's batched algorithm is stuck into a local minimum (but not the converse).
Recently, Hartigan's heuristic~\cite{Hartigan-2013} was suggested to replace Lloyd's heuristic on the
basis that Hartigan's local minima is a subset of Lloyd's optima (Theorem 2.2~of~\cite{Hartigan-2010}).
We argue that this is true only when no {\em Empty Cluster Exceptions} (ECEs) are met by Lloyd's iterations.
Figure~\ref{fig:emptyexception} illustrates a toy data set where Lloyd's $k$-means meets such an empty-cluster exception.
In general at the end of the relocation stage, when points are assigned to their closest current centroids, we may have some empty clusters.
\begin{fact}[empty-cluster exceptions]
Lloyd's batched $k$-means may produce $e=\Omega(k)$ empty cluster exceptions in a round.
\end{fact}
Proof follows from Figure~\ref{fig:emptyexception} by creating $s=n/5$ far apart (non-interacting)   copies of the gadget and setting $k=3s$.

However, those empty-cluster exceptions are a {\em blessing} because we may add $e$ new seeds that will further decrease significantly the cost of $k$-means: This is a partial re-seeding.
Thus the extended Lloyd's heuristic is:
(a) assignment, (b) relocation, and (c) partial reseeding to keep exactly $k$ non-empty clusters for the next stage.
We may use various heuristics for partially re-seeding like the incremental global $k'$-means~\cite{GlobalKmeans-2011} starting from $k'=k-e$ to $k'=k$, etc.

\begin{figure}
\centering
\begin{tabular}{cc}
\begin{minipage}[c]{0.51\columnwidth}
\includegraphics[width=\columnwidth]{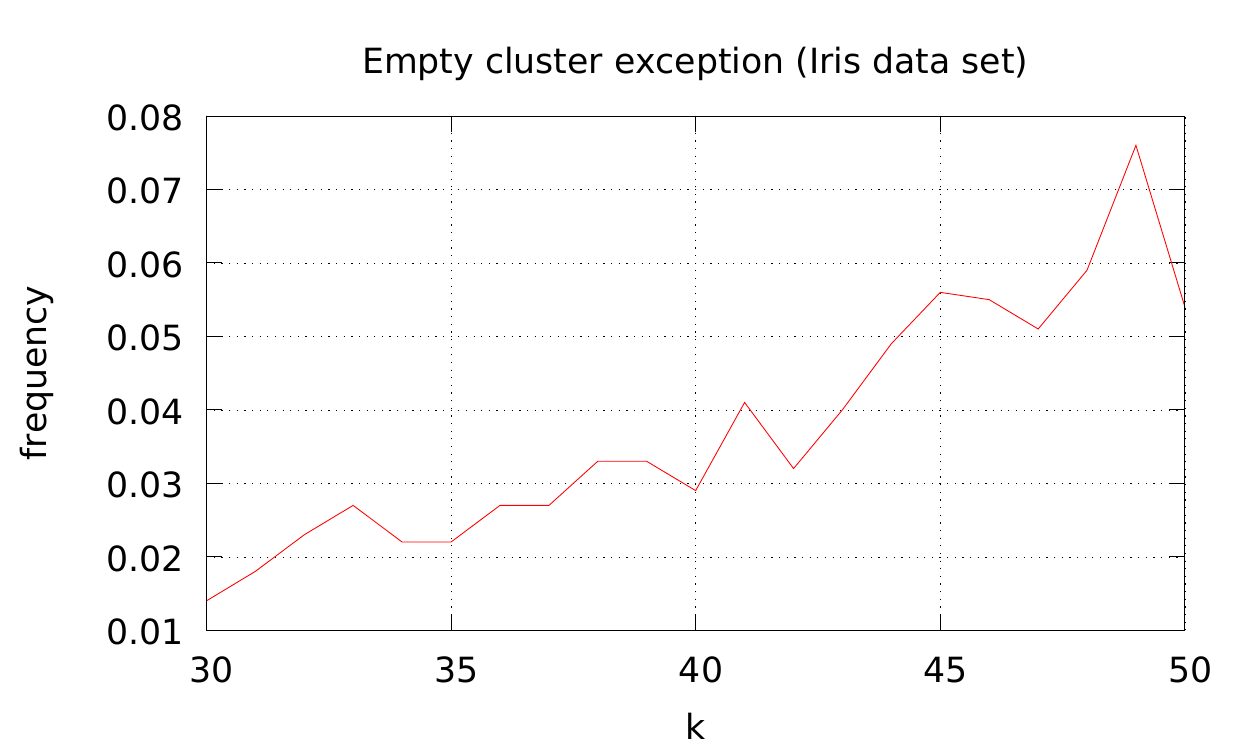}
\end{minipage} &
\begin{minipage}[c]{0.47\columnwidth}
$\begin{array}{|c|ccc|c|}\hline
 & \multicolumn{3}{|c|}{\text{Forgy}} & \text{$k$-means++}\\
\text{$k$} &  e=1 & e=2 & e=3  & e=1   \\ \hline
3 & 48   & &  & 2   \\
5 & 957 & & & 26   \\
10 & 1305 & 8 & 1 &  1   \\
20 & 2718 & 17 & & 98  \\
30 & 12525 & 71  & & 130   \\
40 & 34936 & 382 & & 92   \\
50 & 72193 & 1467 & 7 &  106   \\
\hline
\end{array}$
\end{minipage}
\end{tabular}
\caption{Left: Graph plot of the frequency of empty-cluster exceptions ($e>0$) for Lloyd's $k$-means using Forgy's initialization on the normalized {\tt Iris} data set computed by averaging over a million runs.
Right: Number of ECEs depend on the initialization method: At $k=50$, we observe a frequency of $7.2$\% for one empty cluster, $0.014$\% for two empty clusters, etc. for Forgy's seeding but $k$-means++ initialization produces less such exceptions.
\label{tab:ece}\label{fig:emptycluster}%
}%

\end{figure}

To evaluate the frequency at which those empty-cluster exceptions occur and their number $e$, 
 let us take the {\tt Iris} data set from the UCI repository~\cite{UCI-2007}:
It consists of $n=351$ samples with $d=4$ features (classified into $k=3$ labels) that we first renormalize the data-set so that coordinates on each dimension have zero mean and unit standard deviation.
Let us run Lloyd's $k$-means with (Forgy's) random seed initialization (with a maximum number of $1000$ iterations) for $\mstart=1000000$.
We count the number of empty cluster exceptions and report their frequency in the graph of Figure~\ref{fig:emptycluster}.
We observe that the larger the $k$, and the more frequent the exceptions.
This phenomenon was also noticed in~\cite{KMeansConstrainedCS-2000}. Furthermore, $e$ increases with the dimension $d$ too~\cite{KMeansConstrainedCS-2000}.
However, note that this is a tendency and the number of empty-cluster exceptions vary a lot from a data set to another one (given an initialization heuristic).

%



Let us now run $\mstart=1000000$ $k$-means and report the {\em empirical frequency} of having $e=1, 2, 3, ...$ {\em simultaneous} empty-cluster exceptions.
(Note that our replicated toy data-sets of Figure~\ref{fig:emptyexception} may provide $\Omega(k)$ values).
The empty-cluster frequency depends on the initialization scheme: It is higher when using Forgy's heuristic and lower when using $k$-means++ or global $k$-means. Table~\ref{tab:ece} demonstrates empirically this observation.
As noticed in~\cite{KMeansConstrainedCS-2000}, the number of cluster-empty exceptions rise with $k$ and $d$ and 
the authors~\cite{KMeansConstrainedCS-2000} avoided this problem by setting minimum input size on clusters. 
They surprisingly show empirically that $k$-means with constraints gave better clustering than $k$-means without constraints in practice!


Finally, let us compare the best minimum $k$-means score when performing Lloyd's heuristic (and stopping when we meet an empty cluster exception), and the extended Lloyd's heuristic that partially reseeds the current clustering when the algorithm meets empty-cluster exceptions.
Partial reseeding can be done in many ways by starting from the current number of cluster centers the usual seeding methods (Forgy, $k$-means++ or global $k$-means).
Table~\ref{tab:lloydreseeding} presents the results for the proof of concept using Forgy's re-seeding: 
We  observe that partial reseeding at ECEs allows to reach (slightly) better local minima (see $k=40$ in Table~\ref{tab:lloydreseeding}).

\begin{table}
$$
\begin{array}{|l|cc|cc|c|}\hline
k/\text{method} & \multicolumn{2}{c}{\text{classic Lloyd}} & \multicolumn{2}{c|}{\text{Lloyd+partial reseeding}} & \text{\#ECEs} \\ 
\cline{2-6} 
&  \text{avg} & \text{min} & \text{avg} & \text{min} & \\ \hline 
30 &  12.86 & 9.88 & 12.86 & 9.88 & 7685\\
40 &  9.72 & 7.30 & 9.72 & {\bf 7.28} & 23633\\
50 &  7.5 & 5.47 & 7.5 & {5.47} & 55726\\ \hline
\end{array}
$$
\caption{
\label{tab:lloydreseeding} Comparing Lloyd's $k$-means heuristics without or without partial reseeding (Forgy) when meeting empty-cluster exceptions on {\tt Iris} dataset with a million restart using the same Forgy's initialization at each round. Observe that some better local minima are reached when using partial reseeding at empty-cluster exceptions.}
\end{table}

\section{The blessing of single-point cluster exceptions in Hartigan's heuristic \label{sec:spc}\label{sec:spe}}
Hartigan's heuristic~\cite{Hartigan-2010} consider relocating a single-point provided that it decreases the $k$-means objective function.
In~\cite{Hartigan-2013}, a synthetic noisy data-set is built so that with probability tending to $1$ (as the dimension tends to infinity) 
any initial random partition is stable wrt. Lloyd's $k$-means while Hartigan's converges to the correct solution.
We recall that Hartigan's local minima are a subset of Lloyd's minima~\cite{Hartigan-2010} {\em provided} that Lloyd's heuristic did not encounter empty-cluster exceptions. 
Note that a single-point cluster (with associated cluster having zero variance) cannot be relocated to other clusters since it necessarily increases the $k$-means energy (sum of intra-cluster variances):
\begin{equation}
e(\calP,\calK) = \sum_{j=1}^k v(\calC_j) = \sum_{i=1}^n \|p_i\|^2 - n_j \sum_{j=1}^k  \|c_j\|^2,\ \sum_{j=1}^k n_j=n.
\end{equation}

Table~\ref{tab:SCEs} provides statistics on the Hartigan's $k$-means score and the number of single-point-cluster exceptions (SPCEs) met when performing Hartigan's heuristic.

\begin{table}
$$
\begin{array}{|c|ccc|ccc|}\hline
k & \multicolumn{3}{|c|}{\text{Hartigan's $k$-means}} & \multicolumn{3}{|c|}{\text{Single-point cluster exceptions}} \\ \cline{2-7}
& \text{min} & \text{avg} & \text{max} &\text{min} & \text{avg} & \text{max}\\ \hline
30 & 9.74 & 11.28  & 15.66 & 3 & 20893. 27 & 34007\\
35 & 8.20 & 9.48  & 13.27 & 6 & 43700.20 & 75911\\
40 & 6.98& 8.06  & 12.69 & 12 & 61437.81 & 103407\\
45 & 5.79 & 6.92 & 11.23 & 9 & 83113.54 & 163344\\
50 & 5.06 & 5.95  & 8.96 & 13 & 204222.78 & 367437\\
\hline
\end{array}
$$
\caption{Some statistics on Hartigan's heuristic on the {\tt Iris} data set: min/avg/max $k$-means score and min/avg/max number of single-point cluster exceptions (SPCEs).
\label{tab:SCEs}}
\end{table}

Consider the case of  {\em Single-point-Cluster Exceptions (SCEs)} in Hartigan's scheme where we decide to merge this single-point cluster $\calC_i=\{x\}$ with another cluster $\calC_j$ and redraw another center from $\calP$ (that can thus decrease significantly the variance of the change cluster). 
We accept this relocation iff. this merge\&re-seed operation decreases the $k$-means loss.
For example, when $k=30$ (and $\mstart=1000$), the classical Hartigan's best clustering has $k$-means score $9.75$ while
the heuristic with partial reseeding (associating the single-point clusters to their closest other clusters), we obtain $9.65$.
We keep the experiments short here since the next Section improves Hartigan's heuristic with detailed experiments.


%

\section{A novel heuristic: The merge-and-split-cluster $k$-means\label{sec:merge-split}\label{sec:msc}}
This novel heuristic proceeds by considering pairs of clusters $(\calC_i, \calC_j)$ with corresponding centers $c_i$ and $c_j$.
The {\em basic local search primitive} (pivot) consists in computing the best $k$-means score difference by {\em merging and splitting again} $\calC_{i,j}=\calC_i\cup\calC_j$ with two new centers $c_i'$ and $c_j'$.
Let $\calC_i'$ and $\calC_j'$ denote the {\em Voronoi partition} of $\calC_{i,j}$ induced by $c_i'$ and $c_j'$. 
Since the clusters other than $\calC_i$ and $\calC_j$ are untouched, the difference of the $k$-means score is written as:

\begin{equation}
\Delta(\calC_i,\calC_j) = e_1(\calC_i,c_i)+e_1(\calC_j,c_j)- (e_1(\calC_i',c_i')+e_1(\calC_j',c_j')),
\end{equation}
where $e_1(\calC,c)$ denotes the $1$-means objective function: namely, the cluster variance of $\calC$ with respect to center $c$.
There are several ways (randomized or deterministic) to implement the merge-and-split operation: 
For example, the two new centers can be found by computing:
\begin{itemize}
\item a $2$-means: A brute-force method computes all hyperplanes\footnote{We do not need to compute explicitly the equation of the hyperplane since clockwise/counterclockwise {\em orientation predicates} are used instead. Those predicates rely on computing the sign of a $(d+1)\times(d+1)$ matrix determinant.} passing through $d+1$ (extreme) points and the induced sum of variances of the below-above clusters in $O(n^{d+2})$-time. Using topological sweep~\cite{VarianceClustering-1994}, it can be reduced to $O(n^d)$ time. Note that for $k=2$ and unfixed $d$, $2$-means is NP-hard~\cite{twomeans-nphard-2007}. We can also use coresets to get a $(1+\epsilon)$-approximation of a $2$-means~\cite{kmeans-ptas-2007} in linear time  $O(nd)$.
 
\item a discrete $2$-means: We choose among the $n_{i,j}=n_{i}+n_j$ points of $\calC_{i,j}$ the two best centers (naively implemented in $O(n^3)$). 
This yields a $2$-approximation of $2$-means.

\item a $2$-means++ heuristic: We pick $c_i'$ at random, then pick $c_j'$ randomly according to the normalized distribution of the squared distances of the points in $\calC_{i,j}$ to $c_i'$, see $k$-means++~\cite{kmeanspp-2007}. 
We repeat a given number $\alpha$ of rounds this initialization (say, $\alpha=1+0.01 {n_{i,j} \choose 2}$) and keeps the best one. 
\end{itemize}

When  $\Delta(\calC_i,\calC_j)>0$, we accept replacing $(\calC_i,c_i)$ and $(\calC_j,c_j)$ by $(\calC_i',c_i')$ and $(\calC_j',c_j')$, respectively. Otherwise, we consider another pair of clusters and stop iterating when all pairs do not produce a lower $k$-means score.
This heuristic can be classified as a macro kind of Hartigan-type heuristic that is {\em not} based on local Voronoi assignment.
Indeed, Hartigan's heuristic moves a point $x$ from a cluster $\calC_i$ to a cluster $\calC_j$ and update the two centroids correspondingly.
Our heuristic also change these two clusters but can accept further improvements with respect to a $2$-means operation on $\calC_{i,j}$. 
Thus at the last stage of a Hartigan's heuristic, we can perform this  merge-and-split heuristic to further improve  the clustering.
(This heuristic can further be generalized by simultaneous merging-and-splitting of $r$ clusters.)

\begin{theorem}
The merge-and-split $k$-means heuristic decreases monotonically the objective function and converges after a finite number of iterations.
\end{theorem}
Since each pivot step between Voronoi partitions strictly decreases the $k$-means score $e(\calP,\calK)\geq 0$ by $\Delta(\calC_i,\calC_j)>0$ and that $\min_{\calC_i,\calC_j} \Delta(\calC_i,\calC_j)>0$ is lower bounded, it follows that the 
merge-and-split $k$-means converges after a finite number of iterations.
We compare our heuristic with both Hartigan's ordinary and discrete variants that consists in moving a point to another cluster iff. the two recomputed medoids of the selected clusters yield a better $k$-means score. 
Heuristic performances are compared with the same initialization (Forgy's or $k$-means++ seeding) and by averaging over a number of rounds:
Observe in Table~\ref{tab:msc} that our heuristic (MSC for short) always outperforms discrete Hartigan's method not suprisingly.
Although the number of basic primitives ($\nbop$) is lower for MSC, each such operation is more costly.
Thus MSC $k$-means is overall more time consuming but gets better local optima solutions.
Note that the discrete $2$-means medoid splitting procedure is very well suited for the $k$-modes algorithm~\cite{kModes-1998}, a $k$-means extension working on categorical data sets.

\begin{table}
$$
\begin{array}{|c|l|l|l|l||l|l|}\hline
 \text{Data\ set} & \multicolumn{2}{|c|}{\text{Hartigan}} &  \multicolumn{2}{c|}{\text{Discrete Hartigan}} & \multicolumn{2}{c|}{\text{\bf Merge\&Split}}\\ \cline{2-7}
& \cost & \nbop & \cost & \nbop & \cost & \nbop \\ \hline
\text{Iris} ($d=4$, $n=150$, $k=3$) & 112.35 & 35.11 & 101.69 & 33.54 & {\bf 83.95} & {31.36}\\
\text{Wine} ($d=13$, $n=178$, $k=3$) & 607303 & 97.88 & 593319 & 100.02 & {\bf 570283 }& 100.47\\
\text{Yeast} ($d=$, $n=1484$, $k=10$) &  {47.10}  & 1364.0 & 57.34  & 807.83  & {\bf 50.20}  & 190.58\\
\hline
\end{array}
$$

$$
\begin{array}{|c|l|l|l|l||l|l|}\hline
 \text{Data\ set} & \multicolumn{2}{|c|}{\text{Hartigan++}} &  \multicolumn{2}{c|}{\text{Discrete Hartigan++}} & \multicolumn{2}{c|}{\text{\bf Merge\&Split++}}\\ \cline{2-7}
& \cost & \nbop & \cost & \nbop & \cost & \nbop \\ \hline
\text{Iris} ($d=4$, $n=150$, $k=3$) &  101.49 & 19.40 & 90.48 & 18.93 & {\bf 88.56} & 8.84\\
\text{Wine} ($d=13$, $n=178$, $k=3$) &  3152616 & 18.76 & 2525803 & 24.61 & {\bf 2498107} & 9.67\\
\text{Yeast} ($d=8$, $n=1484$, $k=10$) & 47.41 & 1192.38 & 54.96 & 640.89 & {\bf 51.82} & 66.30 \\
\hline
\end{array}
$$

\caption{Average performance over $1000$ trials of the merge-and-split $k$-means heuristic compared to Hartigan's and discrete Hartigan's heuristics. Top: Common Forgy's initialization and the MSC $k$-means has been implemented using an optimal discrete $2$-means.
Bottom: Common $k$-means++ initialization and the MSC $k$-means has been implemented using a $2$-means++ with $\alpha=0.01\%$.
We observe experimentally that MSC heuristic yields always better performance than Hartigan's discrete single-point relocation heuristic, and is often signigicantly better than Hartigan's heuristic. Note that $k$-means++ seeding performs better than Forgy's seeding
\label{tab:msc}}
\end{table}


\section{Clustering with the $(k,l)$-means objective function\label{sec:klmeans}}

Let us generalize the $k$-means objective function as follows: 
For each data $p_i\in\calP$, we associate $p_i$ to its $l$ nearest cluster centers $\NN_l(p_i;\calK)$ (with $\iNN_l$ denoting the cluster center indexes), and ask to minimize the following $(k,l)$-means objective function (with $1\leq l\leq k$):

\begin{equation}\label{eq:klmeans}
e(\calP,\calK;l) = \sum_{i=1}^n \sum_{a\in\iNN_l(p_i;\calK)}  \|p_i-c_a\|^2.
\end{equation}

When $l=1$, this is exactly the $k$-means objective function of Eq.~\ref{eq:kmeanscost}.
Otherwise the clusters overlap and $|\cup_{j=1}^k \calC_j|=nl$.
Note that when $l=k$, since $\NN_k(p_i;\calK)=\calK$ all cluster centers $c_1, ..., c_k$ coincide to the {\em centroid} $\bar{p}=\frac{1}{n}\sum_i  p_i$ (or barycenter), the {\em center of mass}.
We observe that:
\begin{fact}
$e(\calP,\calK;l) \geq l \times e(\calP,\calK;1)$ with equality reached when $l=k$.
\end{fact}

Both Lloyd's and Hartigan's heuristics can be adapted straightforwardly to this setting.

\begin{theorem}
Lloyd's $(k,l)$-means decreases monotonically the objective function and converge after a finite number of steps.
\end{theorem}

\noindent Proof: Let $c_{2t}$ and $c_{2t+1}$ denote the cost at round $t$, for the assignment ($c_{2t}$) and relocation  ($c_{2t+1}$) stages.
Let $c_0$ be the initial cost (say, from Forgy's initialization of $\calK^0$). For $t>0$, we have:
In the assignment stage $2t$, each {\em point} $p_i$ is assigned to its $l$ nearest neighbor centers $\NN_l(p_i;\calK^{t-1})$.
Therefore, we have $c_{2t} = \sum_{i=1}^n \sum_{c\in \NN_l(p_i;\calK^{t-1})}  D(p_i,c) \leq c_{2t-1}$.
In the relocation stage $2t+1$, each cluster $\calC_j^t$ is updated by taking its centroid $c_j^{t+1}$. Thus we have $c_{2t+1} = \sum_{j=1}^k \sum_{p \in \calC^t_j}  D(p,c^{t+1}_j) \leq  \sum_{j=1}^k \sum_{p \in \calC^t_j} D(p,c^t_j) \leq c_{2t}$.
When  $\calK^{t+1}=\calK^t$ (and thus $c_{2t}=c_{2t-1}$), we stop the batched iterations.

Figure~\ref{fig:klmeans} illustrates a $(k,2)$-means on a toy data-set.

Since $c_t\geq 0$ for all $t$ and the iterations strictly decreases the score function, the algorithm converges.
Moreover, since the number of different cluster sets induced by $(k,l)$ means is upper bounded by $O(n^{kl})$, and that cluster sets cannot be repeated, it follows that $(k,l)$-means converges after a {\em finite number} of iterations.
The bound can further be improved by considering the $l$-order weighted Voronoi diagrams, similarly to~\cite{VarianceClustering-1994}.
Note that the basic Lloyd's $(k,l)$-means may also produce empty-cluster exceptions although those become rarer as $l$ increase (checked experimentally).

\def\ttt{0.15\columnwidth}
\begin{figure}%
\centering
\begin{tabular}{cc}
\fbox{\includegraphics[width=\ttt]{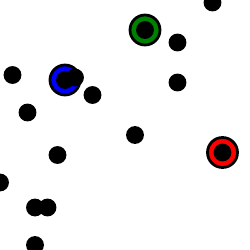}} &\\
\fbox{\includegraphics[width=\ttt]{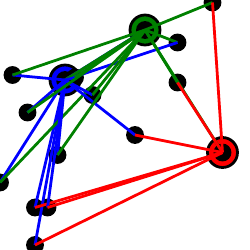}} & \fbox{\includegraphics[width=\ttt]{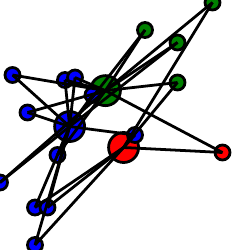}}\\
\fbox{\includegraphics[width=\ttt]{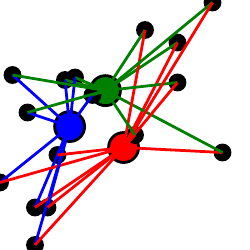}} & \fbox{\includegraphics[width=\ttt]{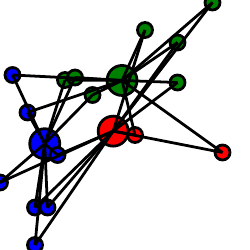}}\\
\fbox{\includegraphics[width=\ttt]{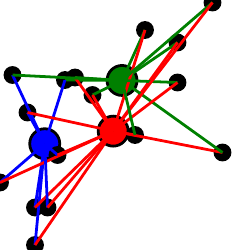}} & \fbox{\includegraphics[width=\ttt]{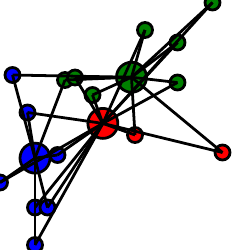}}\\
\fbox{\includegraphics[width=\ttt]{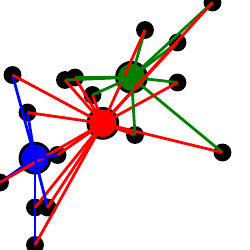}} & \fbox{\includegraphics[width=\ttt]{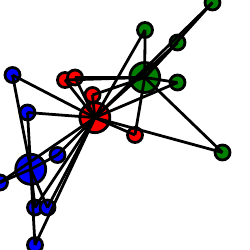}}\\
\fbox{\includegraphics[width=\ttt]{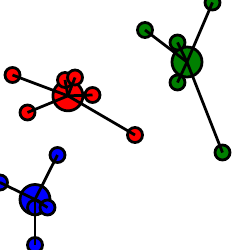}} &  
\end{tabular}
\caption{$(k,2)$-means: Each data point is associated to its {\em two} closest cluster center neighbors. After converging, we relax the $(k,2)$-means solution by keeping only the closest neighbor on the current centroids and run the classic $k$-means. Alternatively, we can relax iteratively the $(k,m)$ means into a $(k,m-1)$-means until we get a $k$-means. \label{fig:klmeans}}%
\end{figure}

Although $(k,l)$-means is interesting in its own (see Discussion in Section~\ref{sec:dis}), it can also be used for $k$-means.
Indeed, instead of running a local search $k$-means heuristic that may be trapped too soon into a ``bad'' local minimum, we prefer to run a $(k,l)$ means for a prescribed $l$. We can then convert a $(k,l)$-means by assigning to each point $p_i$ its closest neighbor (among the $l$ assigned at the end of the $(k,l)$-means), and then compute the centroids and launch a regular Lloyd's $k$-means to finalize:
Let $(k,l)\downarrow$-means denote this conversion.
For example, for $k=6$ and $l=2$, the converted $(k,2)$-means beats the $k$-means $80\%$ of the time for $\mstart=10000$ using Forgy's initialization on {\tt Iris}.
Table~\ref{tab:kvskl} shows experimentally that converted $(k,2)$-means beats on {\em average} the regular $k$-means (for the {\tt Iris} data-set) and this phenomenon increases not surprisingly with $k$. However the best minimum score is often obtained by classical $k$-means. Thus it suggests that $(k,l)$ performs better when the number of restarts is limited. In fact, $(k,l)$-means tend to smooth the $k$-means optimization landscape and produce less local minima but also smooth the best minima.

\begin{table}
{\small
\noindent
\begin{tabular}{ll}
\begin{minipage}[t]{0.45\columnwidth}
$$
\begin{array}{|c|c|c|c|c|c|}\hline
k &  \text{win} & \multicolumn{2}{|c|}{\text{$k$-means}} & \multicolumn{2}{|c|}{\text{$(k,2)\downarrow$-means}}\\ \cline{3-6}
 &   & \text{min} &  \text{avg} & \text{min} & \text{avg} \\
3 & 20.8 & 78.94 & 92.39 & 78.94 & 78.94\\
4 & 24.29 & 57.31 & 63.15 & 57.31 & 70.33\\
5 & 57.76 & 46.53 & 52.88 & 49.74 & 51.10\\
6 & 80.55& 38.93 & 45.60 & 38.93& 41.63\\
7 & 76.67 & 34.18 & 40.00 & 34.29 & 36.85\\
8 & 80.36& 29.87 & 36.05 & 29.87 & 32.52\\
9 & 78.85 & 27.76 & 32.91 & 27.91 & 30.15\\
10 & 79.88& 25.81 & 30.24 & 25.97 & 28.02\\
\hline
\end{array}
$$
\end{minipage} &
\begin{minipage}[t]{0.45\columnwidth}
$$
\begin{array}{|cc||c|cc|cc|}\hline
k & l & \text{win} & \multicolumn{2}{|c|}{\text{$k$-means}}  & \multicolumn{2}{|c|}{\text{$\downarrow(k,l)$-means}}\\ \cline{4-7}
& &  & \text{min} &\text{avg}& \text{min} &\text{avg}\\
5 & 2 & 58.3 & 46.53  & 52.72  & 49.74  & 51.24   \cr
5 & 4 & 62.4 & 46.53  & 52.55  & 49.74   & 49.74  \cr
8 & 2 & 80.8 & 29.87 & 36.40 & 29.87   & 32.54   \cr
8 & 3 & 61.1 & 29.87  & 36.19  & 32.76   & 34.04   \cr
8 & 6 & 55.5 & 29.88  & 36.189  & 32.75   & 35.26  \cr
10 & 2 & 78.8 & 25.81  & 30.61  & 25.97   & 28.23   \cr
10 & 3 & 82.5 & 25.95  & 30.23  & 26.47   & 27.76   \cr
10 & 5 & 64.7 & 25.90  & 30.32  & 26.99   & 28.61   \cr
\hline
\end{array}
$$
\end{minipage}
\end{tabular}

}

\caption{Comparing $k$-means with $(k,2)\downarrow$-means (left) and with $\downarrow(k,l)$-means (right). The percentage of times it outperforms $k$-means is denoted by win. \label{tab:kvskl}}
\end{table}

We can also perform a cascading conversion of $(k,l)$-means to $k$-means:
Once a local minimum is reached for $(k,l)$-means, we initialize a $(k,l-1)$ means by dropping for each point $p_i$ its farthest cluster, perform a Lloyd's $(k,l-1)$-means, and we reiterate this scheme until we get a $(k,1)$-means: An ordinary $k$-means. Let $\downarrow(k,l)$-means denote this scheme.
Table~\ref{tab:kvskl} (right) presents the performance comparisons of a regular Lloyd's $k$-means with a Lloyd's $(k,l\downarrow 1)$-means for various values of $l$ with the initialization of both algorithms  performed by the same seeding for fair comparisons.

%
%

\section{Discussion\label{sec:dis}}
We have extended the classical Lloyd's and Hartigan's heuristics with partial re-seeding and proposed new local heuristics for $k$-means. 
We summarize our contributions as follows: 
First, we showed the {\em blessing} of empty-cluster events in Lloyd's heuristic and of single-point-cluster events in Hartigan's heuristic.
These events happen increasingly when the number of cluster $k$ or the dimension $d$ increase, or when running those heuristics a given number of trials to choose the best solution. 
Second, we proposed a novel {\em merge-and-split-cluster $k$-means heuristic} that improves over Hartigan's heuristic that is currently the {\it de facto} method of choice~\cite{Hartigan-2013}. We showed experimentally that this method brings better $k$-means result at the expense of computational cost.
Third, we generalized the $k$-means objective function to the $(k,l)$-means objective function and show how to directly convert or iteratively relax a $(k,l)$-means heuristic to a $k$-means avoiding potentially being trapped into too many local optima.
$(k,l)$-Means is yet another {\em exploratory} clustering technique for browsing the space of hard clustering partitions.
For example, when $k$-means is trapped, we may consider a $(k,l)$-means to get out of the local minimum and then convert the $(k,l)$-means  to a $k$-means to explore a new (local) minimum. 






\begin{thebibliography}{10}

\bibitem{UCI-2007}
D.J.~Newman A.~Asuncion.
\newblock {UCI} machine learning repository, 2007.

\bibitem{kmeanspp-2007}
D.~Arthur and S.~Vassilvitskii.
\newblock $k$-means++ : the advantages of careful seeding.
\newblock In {\em SODA}, pages 1027 -- 1035, 2007.

\bibitem{bregmankmeans-2005}
A. Banerjee, S. Merugu, I.~S. Dhillon, and J. Ghosh.
\newblock Clustering with {B}regman divergences.
\newblock {\em Journal of Machine Learning Research}, 6:1705--1749, 2005.

\bibitem{ClusteringDynamicProg-1973}
R. Bellman.
\newblock A note on cluster analysis and dynamic programming.
\newblock {\em Mathematical Biosciences}, 18(3-4):311 -- 312, 1973.

\bibitem{KMeansConstrainedCS-2000}
K.~Bennett, Paul~S. Bradley, and Ayhan Demiriz.
\newblock Constrained $k$-means clustering.
\newblock  MSR-TR-2000-65, 2000.

\bibitem{kmeanspp-2014}
A. Bhattacharya, R. Jaiswal, and N. Ailon.
\newblock A tight lower bound instance for $k$-means++ in constant dimension.
\newblock In  {\em Theory and Applications of Models of Computation}, LNCS 8402, pages 7--22, 2014.
%

\bibitem{InitializationKmeans-2012}
S.~Bubeck, M.~Meila, and U.~von Luxburg.
\newblock How the initialization affects the stability of the $k$-means
  algorithm.
\newblock {\em ESAIM: Probability and Statistics}, 16:436--452, 1 2012.

\bibitem{twomeans-nphard-2007}
S. Dasgupta.
\newblock The hardness of $k$-means clustering.
\newblock  CS2007-0890, University of California,  
  USA, 2007.

\bibitem{kmeans-ptas-2007}
D. Feldman, M. Monemizadeh, and C. Sohler.
\newblock A {PTAS} for $k$-means clustering based on weak coresets.
\newblock In {\em SoCG}, pages 11--18.   2007.


\bibitem{Forgy-1965}
E.~W. Forgy.
\newblock Cluster analysis of multivariate data: efficiency vs interpretability
  of classifications.
\newblock {\em Biometrics}, 1965.

\bibitem{Lloydkmeansiterations-2005}
S. Har-Peled and B. Sadri.
\newblock How fast is the $k$-means method?
\newblock In {\em SODA}, pages 877--885. SIAM, 2005.

\bibitem{Hartigan-1975}
J.~A. Hartigan.
\newblock {\em Clustering Algorithms}.
\newblock John Wiley \& Sons, Inc., New York, NY, USA, 99th edition, 1975.

\bibitem{KmeansHartiganWong-1979}
J.~A. Hartigan and M.~A. Wong.
\newblock Algorithm {AS} 136: A $k$-means clustering algorithm.
\newblock {\em Journal of the Royal Statistical Society. Series C}, 28(1):100--108, 1979.


\bibitem{kModes-1998}
Z. Huang.
\newblock Extensions to the $k$-means algorithm for clustering large data sets
  with categorical values.
\newblock {\em Data Min. Knowl. Discov.}, 2(3):283--304, September 1998.

\bibitem{VarianceClustering-1994}
M. Inaba, N. Katoh, and H. Imai.
\newblock Applications of weighted {V}oronoi diagrams and randomization to
  variance-based $k$-clustering.
\newblock In {\em SoCG},  pages 332--339, 1994. 

\bibitem{kmeans-analysis-2000}
T. Kanungo, D.~M. Mount, N.~S. Netanyahu, C. Piatko, R.
  Silverman, and A.~Y Wu.
\newblock The analysis of a simple $k$-means clustering algorithm.
\newblock In {\em SoCG}, pages 100--109.   2000.

\bibitem{DP-kmeans-2012}
B. Kulis and M.~I. Jordan.
\newblock Revisiting $k$-means: New algorithms via {B}ayesian nonparametrics.
\newblock In {\em ICML}, 2012.

\bibitem{GlobalKMeans-2003}
A. Likas, N. Vlassis, and J. J~Verbeek.
\newblock The global $k$-means clustering algorithm.
\newblock {\em Pattern recognition}, 36(2):451--461, 2003.

\bibitem{Lloyd-1957}
S.~P. Lloyd.
\newblock Least squares quantization in {PCM}.
\newblock Technical report, Bell Laboratories, 1957.
\newblock reprinted in IEEE Transactions on Information Theory, March 1982.

\bibitem{MacQueen-1967}
J.~B. Mac{Q}ueen.
\newblock Some methods of classification and analysis of multivariate
  observations.
\newblock  {\em Proceedings Fifth
  Berkeley Symposium on Mathematical Statistics and Probability}, 1967.

\bibitem{kmeansNPhard-2009}
M. Mahajan, P. Nimbhorkar, and K. Varadarajan.
\newblock The planar $k$-means problem is {NP}-hard.
\newblock In {\em WALCOM: Algorithms and Computation}, pages 274--285.
  Springer, 2009.

\bibitem{Xmeans-2000}
D. Pelleg and A.~W. Moore.
\newblock $X$-means: Extending $k$-means with efficient estimation of the
  number of clusters.
\newblock In {\em Proceedings of the Seventeenth International Conference on
  Machine Learning},  pages 727--734, 2000.


\bibitem{Hartigan-2013}
N. Slonim, E. Aharoni, and . Crammer.
\newblock Hartigan's $k$-means versus {L}loyd's $k$-means: Is it time for a
  change?
\newblock In {\em IJCAI}, pages 1677--1684, 2013.

\bibitem{Hartigan-2010}
M. Telgarsky and A. Vattani.
\newblock Hartigan's method: $k$-means clustering without {V}oronoi.
\newblock In {\em International Conference on Artificial Intelligence and
  Statistics}, pages 820--827, 2010.

\bibitem{kmeansexponential-2011}
A. Vattani.
\newblock $k$-means requires exponentially many iterations even in the plane.
\newblock {\em Discrete \& Computational Geometry}, 45(4):596--616, 2011.

\bibitem{GlobalKmeans-2011}
J. Xie, S. Jiang, W. Xie, and X. Gao.
\newblock An efficient global $k$-means clustering algorithm.
\newblock {\em Journal of computers}, 6(2), 2011.

\end{thebibliography}

\end{document}